# VETNET: Assembling Benchmark Datasets for AI-Driven Veterinary Innovation

*A Collaborative White Paper and Proposed Data Ecosystem*

**Katherine J. Koebel**,[1*] Renata Ivanek,[1] Parminder Basran,[2] Jennifer Sun,[3] Casey Cazer,[2,4] Gayatri Anil,[2,4] Tommaso Banzato,[5] Liz Barton,[6] Dottie Cimino Brown,[7] Leszek Bukowski,[8] Marta Castelhano,[2,9] Simon Coghlan,[10] Christopher De Sa,[3] Nikki Dutta,[11] Hala Ghali,[12] Julio Giordano,[13] Joshua Glass,[2] David C. Hogg,[14] Nicholas Hollingshead,[4] Adam Kiehl,[15] Ziang Liu,[3] Monica Meduri,[12] Param Mohapatra,[16] Kim Nayyer,[17] Amelia Nelson,[1] Peter-John Mantyla Noble,[18] Christopher Pinard,[19-22] Beryl Rabindran,[23] Scott Ross,[24] Ron Seccia,[24] Moira Sheehan,[25] Fei Wang,[26] Tracy L. Webb,[15] Kilian Q. Weinberger,[3] Heather M. Whitney,[27] and Chen Xin[2,28]

[*]Corresponding author: kjk239@cornell.edu

1. Department of Population Medicine and Diagnostic Sciences, College of Veterinary Medicine, Cornell University, Ithaca, NY, USA
2. Department of Clinical Sciences, College of Veterinary Medicine, Cornell University, Ithaca, NY, USA
3. Department of Computer Science, Cornell Ann S. Bowers College of Computing and Information Science, Cornell University, Ithaca, NY, USA
4. Department of Public and Ecosystem Health, College of Veterinary Medicine, Cornell University, Ithaca, NY, USA
5. Department of Animal Medicine, Production, and Health, University of Padua, Italy
6. Vet Validate Ltd., St. John's Innovation Centre, Cambridge, UK
7. Mars Veterinary Health, Vancouver, WA, USA
8. VetApp, Poland
9. Cornell Veterinary Biobank, College of Veterinary Medicine, Cornell University, Ithaca, NY, USA
10. School of Computing and Information Systems, The University of Melbourne, Australia
11. Foundation for Food & Agriculture Research, Washington, DC, USA
12. Department of Data Science and Research, The Schwarzman Animal Medical Center, New York, USA
13. Department of Animal Science, College of Agriculture and Life Sciences, Cornell University, Ithaca, NY, USA
14. School of Computer Science, University of Leeds, Leeds, UK
15. Department of Clinical Sciences, College of Veterinary Medicine and Biomedical Sciences, Colorado State University, Fort Collins, CO, USA
16. Center for Applied Mathematics, Cornell University, Ithaca, NY, USA
17. Cornell Law School, Cornell University, Ithaca, NY, USA
18. Small Animal Veterinary Surveillance Network, School of Veterinary Sciences, University of Liverpool, Leahurst Campus, Chester High Road, Neston, Cheshire, UK
19. Department of Oncology, Toronto Animal Cancer Centre, Toronto, Ontario, Canada
20. ANI.ML Research, ANI.ML Health Inc., Toronto, Ontario, Canada
21. Department of Small Animal Clinical Sciences, University of Saskatoon, Saskatoon, Saskatchewan, Canada

22. Centre for Advancing Responsible and Ethical Artificial Intelligence (CARE-AI), University of Guelph, Guelph, Ontario, Canada
23. Amazon Web Services (AWS) Open Data, Arlington, VA, USA
24. Veterinary Medicine Information Technology (VMIT), College of Veterinary Medicine, Cornell University, Ithaca, NY, USA
25. Breeding Insight, Institute for Food and Agricultural Sciences, University of Florida, Gainesville, FL, USA
26. Division of Health Informatics and Artificial Intelligence, Department of Population Health Sciences, Weill Cornell Medicine, New York, NY, USA
27. Department of Radiology, The University of Chicago, Chicago, IL, USA
28. Department of Infectious Diseases and Public Health, Jockey Club College of Veterinary Medicine and Life Sciences, City University of Hong Kong, Hong Kong SAR, China

# Executive Summary

- VETNET represents an ecosystem of live benchmark datasets and evaluation frameworks designed to accelerate AI-driven innovation in animal science and veterinary medicine through standardized, high-quality datasets and multi-institutional collaboration across academic and non-academic sectors.
- This framework addresses the gap between AI advances in human medicine and the fragmented state of veterinary AI development.

Veterinary medicine is entering a period in which artificial intelligence (AI) tools are increasingly accessible to stakeholders (e.g., clinicians, clients, livestock producers, researchers, students, and industry partners) with little to no oversight. Despite this trend, the discipline lacks the requisite infrastructure for supporting the expansion of current, and the innovation of novel, AI capabilities. As a result, the risk that models are developed in a way that is not safe, reliable, clinically meaningful, generalizable, nor sustainable remains a substantial bottleneck for the rate of advancement. We propose VETNET: a collaborative ecosystem for responsible data sharing and AI innovation for animal health and welfare. VETNET will exist not merely as a static repository of datasets, but a coordinated platform for benchmarking resources, task definitions, metadata standards, evaluation metrics, responsible-use guidelines, and sustained stakeholder engagement.

The focus of this effort is to address this need by building an infrastructure at a rate that meets the needs of the profession and improves the standards of products and models that have the potential to impact animal lives. Human medical AI shows that shared data and evaluation can accelerate a field (Whitney and Gruszauskas, 2026), but it also warns against shortcut learning (Banerjee et al., 2023; Hill, Koback, and Schilling, 2024), spurious performance from acquisition or institutional artifacts, distribution shift, dataset drift (Drukker et al., 2023), subgroup failure, weak external validation, and leaderboards that reward narrow optimization without clinical value (Karagyris et al., 2023). VETNET is explicitly designed to create a foundation aimed at avoiding the repetition of these pitfalls while simultaneously accelerating progress.

AI in human medicine provides an example from which the veterinary field may learn and build upon. AI tools are already being adopted in veterinary contexts, including scribes, models for diagnostics, imaging interpretation, client communication, triage, research, and clinical decision assistance. At the same time, veterinary medicine is central to One Health/One Welfare concerns such as antimicrobial resistance, zoonotic disease surveillance, animal welfare, and food safety. This white paper focuses on a defensible evaluation framework rather than treating AI adoption as self-evidently beneficial. It seeks to identify the valid reasons for skepticism surrounding the feasibility, applications, and sustainability of such an infrastructure, then describe in detail the ways in which VETNET will overcome each hurdle.

VETNET's long-term vision of “any animal, any disease, any responsible task” is deliberately broad across the full scope of veterinary medicine. This ambition transcends traditional disciplinary divisions based on species and practice (e.g., primary care, specialty, research). As such, its implementation path must be appropriately structured. The immediate goal is not to solve every available problem, but to first release a narrow, usable benchmark package (i.e., refined data repository, governance structure) that demonstrates VETNET’s ability to deliver upon its intended purpose. From there, the initiative is expected to expand in terms of its membership, data, resources, and practical scope. Once trust and relevance are established through a successfully completed pilot project, VETNET can expand to address the additional needs of the field.

This white paper fundamentally outlines the strategy for moving from vision to implementation. It emphasizes the benchmark dataset philosophy upon which VETNET is built: open where safe, controlled where needed, and federated or hybrid where feasible or preferred. It also identifies the practical components required for success: a candidate pilot project portfolio, governance procedures/documentation, organizational structure, a stakeholder value proposition, measurable milestones including a minimum viable product (MVP), and explicit open questions for the field.

Finally, it should be noted that AI can be used to benefit animals, humans, and the environment, but it can also harm these parties. The enormous environmental costs of model training in terms of energy and water may negate the sustainability benefits of, as an example, improving food animal welfare and productivity with AI tools. Additionally, some AI systems have faced legal scrutiny for violations related to copyright, patient privacy, and intellectual property. As an institution, VETNET acknowledges that some uses, consequences, and processes for AI within animal health are potentially harmful, legally dubious, or ethically controversial. Accordingly, it is crucial that this be accounted for early in its productive life through the involvement of ethicists and legal experts.

# 1 | Current Landscape: Veterinary Medicine in the Age of AI

## 1.1 | The Opportunity

- AI has demonstrated transformative potential in human healthcare and holds the same promise for veterinary medicine, subject to the development of standardized benchmarks and data infrastructure.
- Foundation models, including those in language and vision, as well as developing techniques in interpretable and explainable AI, offer opportunities to revolutionize veterinary diagnostics, treatment planning, scientific research, and clinical decision support.
- VETNET aims to understand the requirements and, if possible, establish this data ecosystem.

To fairly compare the performance of competing algorithms for a given task, machine learning (ML) relies on benchmark datasets: transparent, standardized, and quality-controlled datasets made publicly available for training, testing, and comparative evaluation by a variety of users. They can be thought of as metersticks that ensure the performance of competing models is measured equitably. Benchmark datasets serve five functions: 1) enabling the creation of clinically relevant algorithms; 2) facilitating like-for-like comparison of algorithmic performance; 3) ensuring reproducibility of algorithms; 4) asserting a normative influence on the clinical domains and diversity of patients that will potentially benefit from technological advances; and 5) forming a mechanism for continued validation of models to identify and prevent drift (Panch et al., 2020). These data resources are critical to advancing AI capabilities in the health sciences. Despite their documented importance and the growing demand for AI in animal health applications, the quantity and quality of veterinary benchmark datasets remain notably deficient (Xiao et al., 2025). Veterinary medicine currently lacks the benchmark infrastructure required to evaluate AI tools for practitioners in a manner that is enough to simultaneously engage the AI community. This leaves users vulnerable to systems that suffer from shortcut learning, institutional bias, and poor generalization across species and practice types.

The need is urgent because AI has already entered veterinary contexts through clinical decision support, imaging tools, documentation systems, client-facing chatbots, triage aids, productivity platforms, and general-purpose large language models (Appleby and Basran, 2022). The adoption of generalized AI tools in veterinary contexts can be influenced by biases, terminology, cost considerations, marketing claims, convenience, informal peer use, or human-medicine benchmarks that do not relate cleanly to animal health. These factors are often in conflict with others such as appropriateness, accuracy or value, and wellbeing.

Veterinary medical AI itself presents a unique data science challenge over human counterparts due to the differences in anatomy, physiology, and procedures encountered in animals. Veterinary AI benchmarking differs from human medical benchmarking in several important ways. Models for veterinary tasks must contend with data challenges relating to large inter-/intraspecies variation, client privacy, variable quality of free-text data, different practice management systems, sparse patient-level outcomes data, few negative controls, and uncertainties surrounding data ownership/consent. The absence of a codified ontology as is seen with human medical insurance is a major hurdle (Awaysheh et al., 2017). These differences make direct importation of human medical AI benchmarks inadequate and high risk, even though human medicine offers valuable lessons.

## 1.2 | Convening the Thought Summit

Hosted by Cornell University, the AI Thought Summit entitled “From Data to Animal Health: Building Benchmarks for AI-Driven Veterinary Innovation” was held in Ithaca, New York, on June 9-11, 2026. The four organizers (RI, PB, JJS, and CC) brought together 36 participants, including academics, industry professionals, clinicians, and students. Multi-institutional and transdisciplinary collaboration was emphasized throughout, involving academic researchers, professionals from different fields, and non-academic community stakeholders. Participants from six countries and 16 institutions contributed expertise in veterinary medicine, artificial intelligence, legal and ethical interpretations, and biotechnology data standardization. Degrees held by participants included veterinary doctorates (e.g., DVM, BVSc), board certifications (e.g., DACVR, DACVM, DACVPM), research degrees (e.g., PhD, MS), and other professional graduate degrees (e.g., LLB, MLIS).

The overall purpose of the summit was “to harness AI to transform animal health through collaboration across species and disciplines.” Additionally, four key goals were identified and prioritized:

- Identify key challenges and opportunities in veterinary AI data.
- Build a collaborative, cross-sector stakeholder network.
- Draft a public white paper to guide future efforts.
- Lay the foundation for large-scale, multi-institutional partnerships.

The three-day agenda consisted of three keynote presentations, four panel discussions, two small-group breakout sessions, and nearly ten hours of whole-group discussion. Meeting materials, including notes and presentation materials, were securely maintained and shared with authorized participants in accordance with applicable consent and privacy practices. At the end of each day, student volunteer attendees reviewed all new content and prepared visual (see Appendix A) and plain text thematic summaries, the latter of which formed the basis for this white paper.

The working values derived from the summit were agreed upon as:

- **Animal-centered:** Interpret "for animals" broadly across veterinary contexts while keeping human and environmental connections visible.
- **Transparent and translational:** Support open and ethical research conducted with shared standards that connect rigorous innovation to practical and responsible use cases.
- **Enabling and expeditious:** Facilitate discovery by providing the necessary infrastructure and platform to stimulate collaborative efforts.

**Shared summit framing.** The discussions at the AI Thought Summit converged on VETNET as an effort to create a community initiative that fills the documented need for benchmark dataset infrastructure within veterinary medicine. The agreed grounding language emphasizes animal health and well-being, but also keeps the One Health interconnection among animals, people, and the environment visible, all without making the first implementation target too broad.

## 1.3 | Strategic Positioning

### 1.3.1 | Foundations for Success

*Collaborative Infrastructure*

VETNET's arguably strongest asset is its inclusion of diverse participants whose combined expertise is needed to make the platform a reality. Existing human medical data networks and institutional repositories demonstrate that our endeavor is feasible, albeit challenging. Veterinary medicine is relatively small and more connected compared to human healthcare, which may make coordination easier if governance and incentives are clear. Cloud computing and federated learning architectures also enable secure multi-institutional collaboration. This collaborative infrastructure positions VETNET to achieve success beyond what any single institution or individual could achieve alone. Academic leadership may reduce concerns that benchmark infrastructure is primarily serving proprietary product development, while still actively involving industry participation and expertise.

*Data & Technical Potential*

Veterinary institutions generate vast amounts of underutilized multimodal data (e.g., radiographic images, free-text health records, time-series measurements from sensors). These data span a wide array of contexts that ultimately make the field a data-rich environment inherently well-suited for AI adoption. This is streamlined by the availability of adaptable models and frameworks from the human medical industry. Foundation models and transfer learning enable effective AI development even with smaller datasets than traditional approaches. Furthermore, as AI continues to penetrate society, the resultant digital revolution incentivizes and stimulates the deployment of infrastructure for systematic data collection.

The absence of established veterinary AI data standards and regulation of veterinary medical devices and technologies is a weakness (see below) and an opportunity. These standards do not apply merely to the codification of animal diseases, but to other aspects of datasets including documentation, accessibility, and completeness. Seeing as a model is only as good as the data upon which it is trained, tested, and validated, there is a substantial scientific imperative to ensure datasets used in AI and ML applications conform to agreed-upon industry standards. VETNET can help lead the community to define these standards so they are seen as functional, judicious, and fair. Human health data-sharing efforts prove that governed access to quality-assured datasets catalyzes medical AI research.

*Market & Societal Momentum*

The demand for AI-powered solutions and growing awareness of One Health concerns (e.g., antimicrobial resistance, food safety) create a substantial opportunity for VETNET to address unmet needs in veterinary AI evaluation and benchmarking. In companion animal practice, growing pet ownership and increased expectations and standards for veterinary care are driving demand. As clients increasingly access or request AI-powered tools, the involvement of expert stakeholders, including veterinary informaticians, is necessary to ensure these tools are rigorously developed, evaluated, and deployed.

The stakeholder landscape for veterinary AI tools is notably broader than the initial technical community that creates them. Relevant stakeholders and prospective users include, but are not limited to the following: veterinary colleges and hospitals, general practitioners, specialists, technicians, clients, farm owners and producers, practice management software vendors, diagnostic laboratories, imaging providers, genetic testing companies, insurers, regulators, professional organizations, funders, AI researchers, and commercial AI developers. Animals are the central recipients of veterinary care and are directly affected by decisions supported by AI tools. Their health and welfare interests must therefore be explicitly incorporated into AI development, evaluation, and deployment. Veterinarians have a vital professional and ethical responsibility to represent and advocate for those interests.

### 1.3.2 | Critical Challenges to Address

Because veterinary records lack a universal ontology and span extreme interspecies diversity, VETNET's infrastructure must be capable of mapping local, unstructured clinical notes to interoperable forms rather than relying on pre-existing interoperability.

*Data Complexities*

As alluded to throughout this paper, the lack of universal data standards across veterinary practices and species creates heterogeneity challenges (e.g., interoperability, proprietary data formats). This is further complicated by extreme interspecies, and at times intraspecies, diversity. Individual practice management systems differ in fields, headers, workflows, data exports, and local codebooks. Clinical notes may be rich but inconsistent, and models may learn location-specific terminology and practices and institution-specific documentation habits rather than stable veterinary phenomena.

Relevant starting points may include Digital Imaging and Communications in Medicine (DICOM) for imaging metadata, Systematized Nomenclature of Medicine - Clinical Terms (SNOMED-CT) and veterinary extensions or mappings for terminology, the Observational Medical Outcomes Partnership Common Data Model (OMOP CDM) by Observational Health Data Sciences and Informatics (OHDSI), the Veterinary Breed Ontology for breed representation (Mullen et al., 2025), Health Level Seven (HL7) or Fast Healthcare Interoperability Resources (FHIR)-style lessons from human health data exchange, the Mondo Disease Ontology, and data-card-like schemas for ML-facing dataset documentation (e.g., similar to ATLAS from the Radiological Society of North America). VETNET should evaluate and build on or incorporate these resources rather than waiting for a perfect universal ontology.

*Breadth & Fragmentation*

While the sheer scale of veterinary medicine makes it an exciting field, this may prove to be a challenge for VETNET. Each animal-disease-task combination may require different data, labels, metrics, interpretation thresholds, and validation pathways. A universal taxonomy or ontology will undoubtedly be useful, but anticipating this as a massive, time-consuming undertaking, its absence should not delay progress towards VETNET's initial milestones. Within VETNET, it may be tempting to subdivide participants into many teams based on experience or interests; the resultant teams may be too small to be functional, or lack direction. A defined organizational structure with expectations for active whole-group participation will help minimize fragmentation.

*Resource & Capacity Constraints*

Smaller research funding and market size compared to human medicine limit large-scale dataset development. The result is a smaller sandbox for AI innovation, bounded by limited AI types, labels, metrics, interpretation thresholds, and validation pathways. Active pursuit of funding from multiple sources, including industry, government, and academia, will prove crucial to sustaining the resource needs of a growing VETNET.

### *Ground Truth and Labeling*

In contrast to human medical records, veterinary data is often limited in its ground truth documentation for many conditions due to sparse post-mortem confirmation and diagnostic gold standards. Labels should therefore be treated as designed artifacts, not as simple facts. This raises key questions that must be answered within the documentation of every dataset, including:

- Who produced the label, and under what protocol?
- Was the label generated independently of a given model's developers?
- Does the label represent the same phenomenon that the model is expected to predict?
- Which errors are clinically, economically, ethically, or welfare-relevant?
- Which subgroups, species, breeds, settings, devices, vendors, or geographies require separate performance checks?

### *Trust, Adoption, Bias, & AI Literacy*

Some practitioners may under-trust or be suspicious/skeptical of AI due to lack of knowledge, concerns about environmental impact, black-box behavior, limited evidence, inaccurate outputs, hallucinations, or fear of professional displacement and legal ramifications (Abdelwanis et al., 2024). Others may over-trust AI because the tools are perceived as "smart", sufficiently private, and productivity- and/or efficiency-boosting. VETNET should build trust through transparency and explainability: evidence-based confidence bounded by clear limitations, escalation rules, communication of uncertainty, and domain expert interpretation.

Automation bias and deskilling must be named risks in the use of AI tools. If AI tools are framed as authoritative, clinicians may over-rely on outputs, miss contradictory evidence, or gradually lose practice in core reasoning tasks. Conversely, if AI tools are framed only as threats, useful aids may be rejected even when they improve documentation, access, or safety. VETNET should encourage users of tools developed within its ecosystem to consider not only model performance but also how tools affect human judgment, workflow, learning, and professional responsibility.

### *Equity, Global Access, & Affordability*

Benchmark infrastructure should not require that all users have access to cutting-edge computer resources, highly integrated electronic records, or high-income practice settings. High implementation cost can exclude smaller practices, rural clinics, and low- and middle-income regions or countries, potentially worsening disparities and bias in veterinary AI. VETNET can promote equity through: using user authentication infrastructures (e.g., ORCiD, InCommon) to simplify user access and credentialing; providing lightweight baselines or cloud-based computing resources to ensure models can run on modest computers whenever possible; using standardized input/output schemas that reduce engineering burden for smaller teams; document resource requirements for benchmark participation and model evaluation; designing access tiers that protect sensitive data without making participation impossible for low-resource collaborators; measure performance across practice type, geography, species, breed, equipment, and resource context where data permit; and, leveraging institutional and private sector partnerships to help provide technical staff or infrastructure.

### *Legal Concerns: Privacy, Ownership, Consent, & Liability*

Veterinary medicine lacks a direct HIPAA-equivalent legislation for animal medical records, but that does not make veterinary data ethically, legally, or practically risk-free. Records may contain client information, farm identity, financial details, site metadata, business-sensitive production information, or proprietary vendor data, and are therefore vulnerable to reidentification. Client personal information may also be covered by general data protection regimes such as the General Data Protection Regulation (GDPR) or U.S. state privacy laws, depending on the jurisdiction and data context. Deidentification is necessary but alone insufficient; linkage, inference, model inversion, and misuse remain concerns.

Professional responsibility also remains a critical and central requirement. Good performance

(i.e., as measured by standard AI/ML metrics) of a model developed through VETNET should not imply that veterinarians can delegate responsibility to AI systems. Benchmark performance can support evidence synthesis and candidate selection, but deployment claims require outer-loop validation, appropriate professional oversight, and potentially engagement with regulators or professional bodies. Relevant regulatory examples include the U.S. Food and Drug Administration Center for Veterinary Medicine (FDA-CVM) in the United States, the European Medicines Agency Committee for Veterinary Medicinal Products (EMA-CVMP) in Europe, state veterinary boards, and professional associations.

#### 1.3.3 | Human-Medicine Failure-Mode Lessons to Build Into VETNET

A major lesson from human medical AI is that high retrospective accuracy performance can be misleading. Models may learn shortcuts or employ spurious correlates rather than the intended clinical signal (Drukker et al., 2023; Banerjee et al., 2023; Hill, Koback, and Schilling, 2024). They may also perform well on average while failing in rare but clinically important subgroups and edge cases. These risks are especially relevant in veterinary medicine, where species, breeds, practice settings, equipment, record systems, and workflows vary widely. Accordingly, the following actions should be implemented:

- Use protected and private holdout sets and avoid exposing test labels, potentially through full data sequestration (Baughan et al., 2023).
- Where feasible, include external-site evaluation rather than relying only on random internal splits.
- Use time-based splits when deployment will face temporal change in cases, devices, protocols, or disease prevalence.
- Report subgroup performance across species, breed, age, geography, device/vendor, practice type, timeframe, and data source subsets when sample sizes permit.
- Probe for shortcut features and artifacts that can result through biased metadata or dataset composition, scanner/device signatures, text templates, geographic- and/or institution-specific labels, response variable data leakage, and post-outcome documentation.
- Publish calibration, uncertainty, and abstention behavior for tasks where overconfident errors can cause harm.
- Treat a strong benchmark score as a signal for further validation and not a claim of clinical readiness.
- Define re-validation timeframes.
- Create clear documentation on any data cleaning or preprocessing completed.

Each benchmark should publish not only a score, but also an interpretation guide describing what the score does and does not justify. A high score may indicate a candidate for further validation; it should not be presented as clinical approval. Publishing of model metrics should also be standardized (e.g., AUC, F1 scores) and granular, looking at changes in model metrics across breeds, diseases, geography, time and/or other relevant subsets.

# 2 | VETNET as a Solution

## 2.1 | Vision

VETNET is envisioned as a trustworthy AI ecosystem for the science and practice of animal health, well-being, and veterinary medicine, supporting responsible applications and ethical innovation across its disciplines. In implementation, it will consist of a benchmark dataset repository, wherein each curated dataset is held to quality assurance standards, and a platform that enables equitable, secure access to them for responsible and collaborative model development. VETNET will stimulate the advancement of veterinary AI by enabling the profession to transparently develop, evaluate, and deploy AI tools with high rigor and reproducibility standards before they are used in capacities with clinical, ethical, and legal implications.

The phrase "any animal, any disease, any responsible task" should remain its guiding principle, signaling breadth and inclusiveness simultaneously. However, in practice, this broad goal must be translated into a set of staged, tractable benchmarks. The first year should focus on providing one complete deliverable rather than accumulating many loosely specified datasets.

All individuals and institutions that make up VETNET support one another in ensuring the security and transparency of the data used to create the benchmarks. This support consists of informing all parties, with due discretion, of inconsistencies and errors in the deidentification of the data that form the VETNET benchmarks. Mutual trust and support among all parties are necessary conditions for the success of the entire undertaking.

### 2.2 | Mission

Our mission is to establish VETNET as an accessible, well-governed platform of benchmark resources tailored to the wide array of veterinary-specific use cases. Such resources include the data themselves alongside complete documentation (see Appendix B), metadata, evaluation metrics, holdout sets, and a broader collaborative infrastructure. This will necessitate the development of rigorous, standardized requirements for the quality and content of animal health datasets curated by VETNET. We will achieve this by enacting a governance structure that protects the safety and security of the dataset owners and the entities who appear within it, and by fostering a multidisciplinary network that enables collaboration across all who use or benefit from veterinary AI tools regardless of background. We value AI literacy, equitable participation in scientific advancement, and building trust through transparent communication of the risks, consequences, and far-reaching benefits of AI.

## 3 | VETNET Ecosystem Framework

### 3.1 | Core Components

#### 3.1.1 | Benchmark Repository Infrastructure

- **Sandbox system:** lightweight and preferentially leveraging existing platforms rather than building ground-up
- **Web interface:** integrates uploads, downloads, organizational documentation, and stakeholder communication
- **Tiered access system:** balances open science principles with data privacy, security, and intellectual property concerns (see Table 1 and Appendix C)
- **Rigorous dataset standards:** see 3.1.2 | Standardization
- **Living benchmarks:** versioning, updating, and community contribution mechanisms
- **Tag/filtering system:** comprehensive to enhance user experience in finding datasets (e.g., by species, data format, clinical application)

| Data Sharing/Governance Risk Band | Examples | Suggested Initial Approach |
|---|---|---|
| Low | Deidentified imaging, objective animal-only sensor streams, public research datasets with clean consent and metadata | Possible early pilots if labels, documentation, and clinical-risk disclaimers are adequate |
| Moderate to high | Deidentified clinical records, lab results, structured EHR exports, deidentified notes | Controlled access, strict metadata, documented deidentification, issue-reporting procedures, and explicit restrictions on reuse |
| Very high | Farm identity, production metrics, client financial information, rare cases, proprietary vendor data, cross-border datasets | Controlled access or federated/hybrid evaluation, explicit contributor protections, commercial-use restrictions, legal review, and possibly local computation |

Table 1. Proposed tiers of VETNET benchmark dataset access.

### 3.1.2 | Standardization

- **Metadata:** contextual information about data acquisition/content, DICOM for imaging data
- **Terminology and coding:** conformance to appropriate ontologies to ensure interoperability (e.g., SNOMED CT, Veterinary Breed Ontology)
- **Data deidentification:** uniformly scrub identifiable patient/client/practice information while retaining meaningful information
- **Common data model:** harmonized standards for the content of any uploaded data, metadata, or relevant documentation (e.g., metadata definitions, data types, data processing information)
- **Dataset documentation:** completion of all applicable fields within the common data intake mechanism (see Appendix B)
- **Transparent processing:** generate clear workflow documentations for data cleaning, validation, and preprocessing alongside fully documented code
- **Access and licensing:** users select the access tier in accordance with risk, rights, and intended use cases

### 3.1.3 | Multimodal data integration

- **Clinical imaging:** variety of modalities (e.g., radiography, ultrasound, CT, MRI) and species, potentially including expert annotations
- **Electronic health record data:** available as free text including clinical notes, lab results, treatment protocols, diet, immunizations and vaccine history, and outcomes
- **Time-series data streams:** longitudinal data collected at regular intervals from a given animal, often by sensors
- **Emerging data streams:** other sources of data including behavioral monitoring, genetic/genomic testing, and environmental measurements

### 3.1.4 | Model Evaluation & Validation

- **Evaluation metrics:** industry-standard metrics paired with breed, species, cross species, geography, institution, and out of time subsets and corresponding evaluation metrics that account for veterinary medicine's unique challenges, and create more robust validation pipelines
- **Holdouts/testing mechanisms:** protected subsets to ensure fair model performance

- **Clinical validation:** encourage ethical testing in real-world settings to assess clinical readiness and impact

| Layer | Primary Audience | Overarching Goals | Typical Evidence |
|---|---|---|---|
| Inner-loop benchmark evaluation | AI researchers, benchmark users, technical teams | Evaluates model performance on a defined task under standardized conditions | Hold-out scoring, baseline comparison, calibration, subgroup checks, robustness and sensitivity tests, reproducibility, data leakage checks |
| External and temporal validation | Benchmark stewards, domain experts, model developers | Evaluates areas of bias, and encourages robust creation of hold outs across time, sites, equipment, practices, breeds/species, and documentation styles | External-site splits, time-based splits, domain shift tests, subgroup breakdowns, artifact/shortcut probes, model drift |
| Outer-loop veterinary validation | Clinicians, hospitals, farms, regulators, professional bodies, clients, producers | Evaluates and encourages improvements and supports safety in real-world applications in relation to welfare, access, and cost outcomes | Domain expert thresholds, prospective testing, human-factors studies, workflow fit, safety monitoring, real-world outcome analysis |
| Governance and claims review | Contributors, institutions, industry, legal/ethics bodies | Assess the reliability and validity of claims for access, marketing, and reuse | License terms, data-use agreements, audit logs, misuse response, public/private reporting rules |

Table 2. Components of model evaluation and validation in the VETNET framework.

## 3.2 Governance & Sustainability

### 3.2.1 | Organizational Structure

- **Multi-stakeholder involvement:** across professional domains (e.g., clinical practice, livestock production, research, industry), expertise (e.g., ML, veterinary medicine), and focus (e.g., small animal, livestock, imaging, surgery)
- **Adapt existing structure from human health equivalent:** consider repurposing the internal governance of a human health initiative such as the Medical Imaging and Data Resource Center (MIDRC; https://www.midrc.org) to serve as a blueprint
- **Working groups:** four teams (Minimum Viable Product (MVP), Governance, Stakeholders and Funding, and Repositories), each focused on a single critical aspect of VETNET's continued success (see below)
- **Documentation:** bylaws for official member expectations and collaboratively written standards for all curated data
- **Interest groups:** connect contributors based on expertise or interests
- **Independent vs institutional:** decide whether VETNET should be a free-standing entity or housed within an existing institution, likely within academia

### *Minimum Viable Product (MVP) Working Group Statement*

The assembly and deployment of a complex multi-site healthcare benchmark dataset is a challenging process that has yet to be thoroughly described in the veterinary space. There is a need for standard operating procedures for the implementation of each layer of a data standards and interoperability stack that supports dataset development. The MVP Working Group brings together professionals with diverse areas of expertise and the ability to establish such procedures. In collaboration with other working groups, it will focus on the mechanics of data collection, harmonization, and deployment, building towards a deployed, small-scale benchmark dataset. This proof-of-concept-focused work will aim to establish a foundational model for future VETNET dataset curation, identify gaps in the current proposed plans and available resources, and demonstrate the feasibility of multi-site collaboration.

### *Governance Working Group Statement*

Governance spans both the internal processes of VETNET as an organization, as well as the standards to which curated datasets are held. Animals' interests should always be considered through medical/welfare expertise and explicit consideration of clinical outcomes. Furthermore, VETNET should not ask contributors simply to donate data alone. It should offer contributors an onboarding pathway, predictable governance, credit, collaboration, protection, and a path from real veterinary problems to usable AI benchmarks. For agricultural and farm animal production data, a contributor-rights framework could include principles similar to those in the Farmers' Bill of Rights. Additional consideration should be given to veterinary clients and others who share data about their animals and/or themselves and who thereby also take some risk that such information could be misused or revealed (e.g., through re-identification). This includes consideration of the circumstances of consent to the initial use and to the potentially subsequent and additional employment of that data. A principle of informed consent from data ‘owners’ for recording and using certain data will be relevant here.

| Governance Mechanism | Questions to Answer |
|---|---|
| Access review | Who can access each dataset? What credentials, agreements, or approvals are required? How long is access granted? |
| Revocable access | Can access be removed after misuse, role change, or project completion? Who decides? |
| Contributor withdrawal/correction | What happens if a contributor discovers an error or requests correction or withdrawal? |
| Versioning and deprecation | How are benchmark changes documented and communicated? How are old scores handled after dataset, terminology, or metric updates? |
| Protected test sets | Who can see the test labels? How are submissions limited? How is leakage detected? |
| Commercial-use rules | Can companies train on the data? Can they market benchmark scores? What disclaimers are required? |
| Misuse response | What happens if a model is marketed as clinically or practically validated based only on benchmark performance? |
| Attribution and benefit sharing | How are contributors credited? What citation, authorship, or acknowledgment language is required? |
| Audit and logging | What access, downloads, submissions, and scoring events are logged? Who reviews them? |
| Data quality escalation | Who adjudicates label disputes, missing metadata, suspected bias, or discovered leakage? |
| Maintenance | How often are the datasets and platform reviewed and validated? How often are the governance protocols reviewed and updated? |

Table 3. Considerations of VETNET governing participants.

### *Stakeholders and Funding Working Group Statement*

The VETNET Stakeholders and Funding Working Group aims to secure the funding, partnerships, and institutional commitments needed to launch and sustain a trusted resource for veterinary AI benchmarking and research. Its work centers on developing a diversified funding strategy spanning grants, industry partnerships, institutional support, philanthropy, and potential evaluation services. The group's proposed two-year goals include securing funding, establishing 5–7 institutional data partnerships, and attracting 10–15 industry consortium members, while funding the personnel and infrastructure required for continued operation.

The committee's activities include estimating VETNET's minimum operating costs, identifying anchor funders and founding partners, developing partnership packages, and defining transparent rights and responsibilities for commercial participation. It also supports access to existing datasets by developing contributor incentives, attribution, protections, and benefit-sharing arrangements. These efforts seek to balance open science with privacy and commercial constraints, preserve independent evaluation, and ensure that participation benefits institutions and contributors across different resource settings. Early discussions include whether industry should pay for benchmark access, evaluation services, or both.

A current focus is applying for grants, such as the NSF AI Datasets initiative and other international granting agencies, that would improve the usability of existing veterinary data through harmonization, metadata generation, linkage, and reusable AI-assisted workflows. Such initiatives require input from data stakeholders on available datasets, potential institutional partners, existing standards and infrastructure, scientific use cases, and governance and international collaboration requirements. This input will help refine the concept's scope, clarify its technical contribution, and establish credible deliverables.

### *Repositories Working Group Statement*

The goal of the Repository Working Group is to create the architecture and infrastructure of the VETNET platform, establishing the technical foundation needed to securely, consistently, and efficiently integrate datasets into the repository and utilize them. This goal includes data infrastructure, clear and consistent data documentation, and user-friendly interfaces.

- **Architectural design**
  - Determine where and how the VETNET platform will be hosted and deployed for the prototype and long-term solution.
  - Define the overarching data architecture as well as any additional ingression and egression steps.
  - Design the overall repository structure to support diverse datasets and future scalability.
- **Computational design and data ingression** (in collaboration with the MVP team)
  - Develop a clear ingression pipeline that guides dataset owners through the process of preparing data documentation and contributing data to VETNET. The pipeline will include:

- Automatic deidentification of sensitive data while keeping relative information and associated metadata
  - Automatic standardization of data formatting and terminology
  - Semi-automatic metadata generation at both the database and file levels
  - Automatic validation checks to identify errors, inconsistencies, or missing information before data enters the repository
- Define the Common Data Model: establishing metadata and structural requirements needed across VETNET while accounting for the diversity of datasets, formats, and intended uses. Key questions will include what format best supports the repository architecture and how the platform can maintain consistency without unnecessarily restricting the types of data that can be contributed. This includes:
  - Database-level metadata requirements
  - File-level metadata requirements
  - Standard formats and structures that account for diverse data types and use cases
  - Relational databases linking datasets together as well as secured raw portions of the data
  - A datalake or datahouse architecture that balances consistency with the varied needs of VETNET datasets

- **Communication, documentation, and user interface**

We must develop clear documentation and user-facing infrastructure to make the repository easy to understand and use, while helping dataset owners make informed decisions about how their data can be shared and used within VETNET. This includes:

- Documentation and guidance for dataset owners throughout the ingression process
- Clear expectations of repository requirements and workflows
- Clear documentation for dataset licensing options, and communication to the dataset owners through displaying the differences, requirements, and implications of each license
- User interfaces and communication tools that make technical requirements easier to navigate

### 3.2.2 | Ethical Guidelines & Responsible AI

- **Bias assessment:** systematically test for model/data flaws such as shortcut learning or underrepresentation
- **FAIR guiding principles:** Findable, Accessible, Interoperable, and Reusable (Wilkinson et al., 2016)
- **Animal welfare:** conduct impact assessments, including considerations of animal autonomy and quality of life
- **Consultation with ethicists:** obtain professional guidance on ethical implementation and oversight
- **AI literacy:** must ensure VETNET packages and their products are explainable and understandable by users, especially clinicians, potentially by directing users to training resources
- **Environmental sustainability:** consider the energy and water usage of data and models curated, hosted, and made available through VETNET
- **Incentivization:** encourage data sharing from those who originate or legitimately possess data

### 3.2.3 | Legal & Regulatory Alignment

- **Proactive engagement:** involve domestic and international regulatory bodies to shape AI approval pathways

- **Governance policies:** documented standards and guidelines addressing ownership, consent, privacy, and secondary use rights
- **Liability:** appropriately allocate responsibility between dataset donors, veterinarian users, individuals, and institutions
- **Legal counsel:** involve law professionals/experts in VETNET early to obtain guidance

## 3.3 | Minimum Viable VETNET: A Proof-of-Concept Benchmark Dataset

### 3.3.1 | Components

The first VETNET deliverable should be a deployed, small-scale benchmark dataset. It should be narrow enough in scope to successfully launch but complete enough for both the veterinary and AI communities to assess and critique it. This project will examine the mechanics of benchmark dataset deployment as defined in the VETNET data standards and interoperability stack and expose knowledge and resource gaps that can be addressed by future larger-scale efforts. The MVP Working Group emphasizes the importance of this proof-of-concept as a study in "process over product" that will serve as a foundation for future efforts.

| **Component** | **Questions to Answer** |
|---|---|
| Task definition | What problem does this solve within veterinary medicine? Who are the intended users? What are the inputs and outputs? |
| Governance terms | How are data controlled? Who will have access? |
| Data and metadata retrieval | How are data obtained? What metadata are included? (*Collaboration with Repository Working Group*) |
| Annotation and coding | How are data labeled/annotated/tagged in a uniform way across sites? What standardization, if any, is applied to the terminology? (*Collaboration with Repository Working Group*) |
| Data deidentification | What steps are taken to scrub identifiable information? What potentially identifiable information remains in the dataset? (*Collaboration with Repository Working Group*) |
| Common data framework | Do the data conform to an agreed-upon format? (*Collaboration with the Repository Working Group*) |
| Data exchange | How are data securely transmitted? |
| Dataset documentation | Are the contents thoroughly and transparently explained/defined? Does the current documentation enable an outside user to fully understand its contents and use? (*Collaboration with the Repository Working Group*) |
| Access control and licensing | Who gets to see the data? How can they use it? |
| Dataset deployment | How are the data made available and delivered to users? |

Table 4. Components of the VETNET data standards and interoperability stack.

### 3.3.2 | Candidates for Pilot Projects

The first benchmark datasets should be chosen through community priorities determined through engagement, not by data availability alone. Criteria should include stakeholder demand, availability of quality data, risk-benefit, clinical or welfare value, ML interest, equity implications, and ability to define expert thresholds.

| Candidate | Why It Belongs | Risks and Considerations |
|---|---|---|
| Medical imaging (*Frontrunner*) | Clear input structure; DICOM precedent; strong ML interest; specialist-access problem; likely easier to document than free text | Diagnostic stakes are high; labels need expert review; benchmark results may be misused in marketing; strong first candidate if tightly scoped and framed as research benchmark, not clinical approval |
| AI scribes/documentation tools | Already entering practice; clear practitioner pain point; useful for benchmarking hallucination, omission, completeness, privacy, and record usability | Sensitive conversations and medical-record integrity concerns, as well as the potential to steer clinical care in unintended ways; good early, controlled product evaluation benchmark or challenge (Francis, Whitney, and Shakhsheer, 2026) |
| Client-facing chatbots (e.g., triage) | Clients already use public AI tools; can test safety, escalation, uncertainty, and inappropriate reassurance | High risk if outputs delay care; useful if framed around safe escalation, long-term client communication and retention (i.e., for followups and yearly care) and bounded guidance rather than diagnosis |
| Agentic clinical workflows | Captures emerging use of LLMs for multi-step reasoning, follow-up questions, summarization, and decision support | Hard to evaluate; may require simulated cases and expert rubrics; important medium-term benchmark unless tightly scoped |
| Wearables/animal-state prediction | Objective sensor streams; welfare relevance; time-series ML fit; companion, equine, and production relevance | Vendor-specific devices and farm/commercial sensitivity; strong candidate if metadata and device documentation are rigorous and deidentified |
| Pain/welfare assessment | High ethical value; aligns VETNET with welfare, not only health | Pain is inferred and label uncertainty is substantial; useful if expert protocols and threshold language are strong; substantial risk of |

| Candidate | Why It Belongs | Risks and Considerations |
|---|---|---|
| | | inappropriately and inadequately measuring pain and welfare |
| Coding/ontology mapping from records | Addresses the standardization bottleneck and can structure unstructured legacy records; can also help to create a structured layer over the raw data layer for future AI systems | May learn institution-specific documentation habits and lack generalizability; good infrastructure-enabling benchmark |
| Commercial product evaluation | Addresses current market opacity and practitioner trust | Requires mature legal and governance rules for public/private reporting, claims, and benchmark gaming |

Table 5. Candidates for the pilot VETNET benchmark dataset.

# 4 | VETNET Implementation Roadmap

## 4.1 | Phase 1: Foundation (Years 1-2)

The first year (see Table 6) will involve a series of quarterly goals. The overarching milestone is for VETNET to produce one complete benchmark dataset loop: choose a task, secure data or controlled access, define labels and thresholds, build a runnable harness, release documentation, collect feedback, and revise.

VETNET has the following goals for Phase 1:

- Establish VETNET governance structure (see Table 7), secure founding partnerships, and initial funding commitments.
- Develop and pilot up to 3 benchmark datasets in high-impact areas (e.g., canine/feline radiography, companion animal oncology, livestock disease surveillance).
- Create standardized protocols for data collection, annotation, quality control, and benchmark evaluation metrics.

| Period | Milestones | Primary Output(s) |
|---|---|---|
| 0-3 months | Confirm organizational identity (e.g., mission, vision, executives); assemble interest groups; screen existing domain-specific databases; identify candidate pilot projects; appoint project champion; convene four Working Groups; identify data contributors | White paper publication; public-facing Website; Working Group launch plans; preliminary organizational structure |
| 3-6 months | Draft access and commercial-use language; create metadata/readme/data-card templates; define annotation protocol; identify protected holdout approach; document data provenance and cleaning requirements; deploy an early prototype with limited capabilities | Benchmark documentation and governance template package as well as a prototype |
| 6-9 months | Build the evaluation harness; implement baseline model or baseline workflow; finalize scoring scripts; define expert threshold worksheet; test submission or scoring process internally; deploy advanced capabilities and automated ingression pipelines | Internal runnable MVP with baseline and threshold guide |
| 9-12 months | Release or controlled-launch pilot benchmark datasets; collect user feedback; resolve issues; document limitations; decide whether the task is ready for broader release, revision, or outer-loop validation | Pilot benchmark dataset release(s) and lessons-learned report; peer-reviewed publication in major journal to publicize the success of our demo use case |

Table 6. Quarterly milestones within Year 1.

| Role/Capacity | Reason for Need | Phase 1 Responsibility |
|---|---|---|
| Executives/Board of Directors | Prevents the initiative from becoming another broad planning exercise with no deliverable | Own timeline, convene decisions, assign and monitor progress on tasks, resolve scope conflicts, maintain momentum |
| Technical coordinator | A runnable benchmark requires code, documentation, packaging, scoring, and issue resolution | Build and maintain the framework, repository structure, versioning, and submission workflow |
| Data steward/curator | Benchmark quality depends on ingestion, provenance, cleaning, metadata, and QA | Coordinate dataset templates, quality checks, cleaning logs, and documentation |
| Domain expert reviewers | Labels, metrics, and thresholds must be veterinary meaningful to the veterinary field | Define task, labels, error severity, thresholds, and interpretation guide |
| Legal/ethics leads | Data use, consent, liability, commercial claims, and contributor rights require proactive policy | Draft access, attribution, commercial-use, privacy, and consent language |
| Community manager/coordinator | Sustained participation requires communication, onboarding, and follow-up | Run listserv/community channel, collect feedback, manage working groups |

| Role/Capacity | Reason for Need | Phase 1 Responsibility |
|---|---|---|
| Infrastructure support | Controlled access, hosting, security, cloud costs, and authentication cannot be *ad hoc* | Select platform, implement access controls, manage storage/hardware, monitor metrics and security. |
| Funding/admin support | Long-term maintenance cannot rely only on volunteer enthusiasm | Identify grants, institutional support, cloud credits, philanthropy, and industry support; develop marketing strategy and monitor use metrics (or separate marketing personnel) |

Table 7. Key roles within the proposed VETNET internal governance.

## 4.2 | Phase 2: Expansion (Years 2-3)

After the first benchmark dataset loop is demonstrated, VETNET should expand cautiously and strategically. Expansion should prioritize stronger standards, new data types, improved documentation, additional modalities, more contributors, clearer thresholds, and a more complete governance structure rather than simply increasing the number of datasets.

VETNET has the following goals for Phase 2:

- Scale from the initial benchmark(s) to a small portfolio across species, modalities, and clinical domains only after the MVP workflow is maintainable.
- Launch VETNET certification program for AI tools meeting benchmark performance and safety standards.
- Establish international data-sharing agreements and contributor workflows based on pilot experience.
- Explore hybrid or federated learning infrastructure.
- Develop public/private reporting rules for product evaluation and benchmark claims.
- Create educational materials for practitioners explaining what benchmark scores mean and what they do not mean.
- Develop low-resource participation pathways, including lightweight baselines, simplified submission formats, and documented compute requirements.
- Develop marketing and outreach strategies and use/impact metrics.
- Develop and implement validation and review protocols.

## 4.3 | Phase 3: Maturation (Years 4+)

VETNET has the following goals for Phase 3:

- Achieve a self-sustaining operational model through certification fees, licensing, and industry partnerships.
- Demonstrate measurable clinical impact through prospective studies of benchmark-validated AI tools.
- Influence regulatory policy and establish VETNET standards as global reference for veterinary AI.

### 4.4 | Summary of Roadmap

| Phase | Governance Needed | Pitfalls to Avoid |
|---|---|---|
| Foundation | Project champion; small steering group; Working Group leads; one technical maintainer; one legal/ethical consultant; website/platform development to improve visibility and start building engagement; clear escalation rules | Creating many committees before a benchmark exists; unresolved scope debates; governance processes that contributors cannot understand |
| Expansion | Formal steering committee; data governance board; defined technical standards; legal/ethical counsel; domain-specific interest groups; marketing and website/platform maintenance personnel; contributor representation; unified messaging to drive engagement and build strong reputation | Scaling datasets faster than governance, documentation, or maintainer capacity; lack of reach |
| Maturation | Sustained governance with rotating membership; transparent policies; audit mechanisms; strategic messaging to maintain engagement and sustain reputation; external advisory input; validation/certification discussion only if evidence supports it | Treating benchmark maintenance as a one-time project; allowing commercial claims to outrun evidence |

Table 8. Phases of VETNET's 5-year roadmap.

## 5 | Anticipating Success: Factors and Metrics

Success requires low-friction access where appropriate, clear documentation, visible early wins/measurable impact, regular communication, responsive issue resolution, and stakeholder-specific value propositions. It may be assessed through quantifications relating to VETNET's deliverables, including the successful deployment of high-performing benchmarks, peer-reviewed publications, and the population of animals whose care is influenced by VETNET. Other, more subjective impacts, such as the net contribution to animal health/welfare and increased standards of veterinary care, are less quantifiable but equally important.

The success of VETNET will depend on a collection of interrelated factors that will cumulatively grow the initiative from its early stages into a long-standing institution within the field.

### 5.1 | Community Engagement and Usability

- Active participation from a variety of stakeholders within the domain of veterinary AI (see Table 9)
- Transparent communication about AI capabilities and limitations building trust and appropriate adoption
- Training programs developing AI literacy across the veterinary workforce
- Equitable access with user-friendly interface and data contribution processes

| Stakeholder | Contributions | VETNET value proposition |
|---|---|---|
| Veterinary colleges and hospitals | Clinical datasets, specialists, trainees, validation sites | Credit, shared infrastructure, benchmark leadership, research visibility, student training and AI literacy resources, and improved validation pathways |
| Veterinary general practitioners | Real-world need, workflow knowledge, broad case mix, adoption feedback | Practical guidance, trustworthy evidence, AI literacy resources, clinically meaningful thresholds |
| Veterinary specialists | Expert labels, task definitions, gold-standard review, error-severity judgment | Influence on standards, protection against weak tools, evidence for appropriate use |
| Companion animal clients | Consent, history, owner-observed outcomes, privacy expectations. | Improved validated tools and clearer communication about AI utility and limitations |
| Farm owners/producers | Production data, herd/flock context, operational constraints, privacy expectations | Contributor protections, controlled access, benefit sharing, tools for productivity, welfare, and disease surveillance |
| Practice-management-system vendors and diagnostic providers | Data infrastructure, record formats, exports, images, lab data | Standards that reduce friction and create credible product ecosystem |
| Industry AI developers | Models, tools, infrastructure, funding | Credible comparison, reproducible validation, path from benchmark to clinical evidence |
| Regulatory/professional bodies | Guidance, standards, professional expectations, access to practitioners | Evidence base for staged guidance without replacing regulatory review |
| AI researchers and trainees | Methods, baselines, evaluation code, reproducibility culture, technical support/workforce | Usable tasks, leaderboards, baselines, publication value, and domain relevance |
| Low-resource and global partners | Diverse deployment contexts, disease and production settings, equity priorities | Participation pathways that do not require high compute or elite institutional infrastructure |

Table 9. Value map for VETNET stakeholders.

## 5.2 | Quality, Rigor, & Maintenance

- Expert curation and multi-reviewer consensus for ground truth labels
- Diverse representation across breeds, species, practice settings, and geographic regions
- Continuous quality monitoring and benchmark refreshing by dedicated groups/individuals to prevent dataset drift

## 5.3 | Impact Measurement

- Track clinical outcomes, not just technical performance metrics
- Assess objective and subjective indicators of success

- Monitor equity, access, and environmental implications of AI deployment
- Demonstrate value proposition for multiple stakeholders

| Category | Example Measures |
|---|---|
| Inner-loop usage | Submissions, active teams, reproducible baselines, issue resolution, benchmark citations, performance improvements, publications/citations |
| Repository health | Versioned releases, maintenance cadence, documentation completeness, metadata quality, access review time, protected holdout integrity |
| Failure-mode control | External-site performance, time-based validation, subgroup breakdowns, calibration, artifact/shortcut probe results, leakage reports |
| Stakeholder trust | Contributor retention, practitioner feedback, user sentiment and referrals |
| Engagement metrics | Industry participation under transparent rules, professional-body engagement, Web traffic, downloads/uploads, community forum discussion, anecdotes of clinical implementations |
| Outer-loop impact | Prospective validation, workflow fit, safety outcomes, welfare effects, human-factors implications, standard-of-care analysis, real-world monitoring |
| Equity, access, and sustainability | Participation from smaller practices and lower-resource settings, compute requirements, geographic representation, performance across resource contexts |

Table 10. Categorization of success evaluation areas for VETNET.

# 6 | Call to Action

VETNET represents a transformative opportunity to position veterinary medicine at the forefront of responsible AI innovation. Success requires:

- **Institutional commitment**: veterinary colleges, companies, and practices dedicating resources and expertise to benchmark development
- **Industry partnership**: AI companies and technology providers engaging collaboratively rather than competitively
- **Regulatory collaboration**: proactive dialogue with governmental and international bodies shaping enabling policy frameworks
- **Philanthropic investment**: foundations as funding agencies recognizing animal health AI as a societal good with appropriate benchmarks and validation

We welcome, encourage, and appreciate contributions to our initiative in any of the following capacities:

- **Veterinary institutions:** nominate pilot tasks, identify candidate datasets, provide domain experts, and help define clinically meaningful thresholds
- **AI researchers and technical teams:** help build the evaluation harness, baseline models, input/output schemas, scoring scripts, reproducibility checks, and failure-mode probes
- **Industry partners:** participate collaboratively rather than fragmenting the field; contribute data, tools, infrastructure, or funding under transparent benchmark and commercial-use rules
- **Regulatory/professional bodies and legal/ethics experts:** clarify appropriate-use expectations, claims boundaries, consent language, PII issues, and staged validation pathways
- **Funding agencies and philanthropic organizations:** support building and maintaining infrastructure for data storage, organization, security, and management, data stewardship, technical maintenance, governance, global participation, and community infrastructure, not only model development
- **Clients and animal caregivers:** identify real workflow and welfare pain points and participate under governance rules that protect privacy, credit, and legitimate commercial interests; remain active in discussions on the use of your animals' data and familiarize yourself with AI literacy

# 7 | Open Decisions for Summit Participants

Many questions remain about the current and future status of VETNET. We are actively working to answer them, but also solicit outside input towards their resolution.

| Question | Why It Matters | Concrete Output Needed |
|---|---|---|
| What are VETNET's institutional identity (e.g., mission, vision, values)? | Discussion converged on responsible data and AI innovation for animals, grounded in animal health and well-being, veterinary science, and One Health interconnections | Comments welcome on the current mission/vision language and a concise values list |
| What is VETNET at launch? | The initiative could be a benchmark platform, data catalog, governance framework, evaluation authority, community network, or some combination | One-sentence launch definition and explicit non-goals for year one |
| Which first pilot should be selected? | The first benchmark must be feasible, useful, and trust-building; a poor first task can stall the initiative | Comments welcome on the ranked shortlist made using feasibility-risk-value criteria; named owner for the top pilot |
| What access model should the pilot use? | Open, registered, controlled, and federated access have different implications for participation, privacy, and contributor trust | Pilot-specific access tier and draft data-use rules |
| How should companies participate? | Industry is essential, but commercial use, benchmark claims, product evaluation, and derivative products require clear boundaries | Commercial-use and benchmark-claim policy for the pilot |
| What expert threshold defines success? | ML metrics alone cannot define clinical or welfare meaning | Threshold worksheet completed by domain experts before release |
| How should animal welfare interests be represented? | Animals are the subjects affected by benchmarked systems but cannot verbally consent or advocate | Welfare review expectations and minimum welfare-relevant outcomes for relevant tasks |
| How will equitable global and low-resource access be protected? | High compute and implementation cost can concentrate benefits in wealthy practices and countries and lack important data and insight | Equity design requirements for data access, baselines, documentation, and participation |
| Who maintains VETNET after the summit? | Without named capacity, VETNET risks becoming another planning document | Named project champion, technical maintainer, data steward, and next milestone |
| How do we communicate VETNET to potential participants? | Transparency, understanding and engagement | Communications group to determine messaging, channels, tactics, and timeline |

Table 11. Unanswered questions regarding the future of VETNET.

# 8 | Conclusion

VETNET seeks to establish the evidence base and infrastructure necessary for veterinary medicine to harness AI's transformative potential responsibly. By learning from human medicine's successes and failures, veterinary medicine can establish benchmarks that are rigorous, equitable, interpretable, and clinically impactful from the outset.

VETNET represents a chance for veterinary medical professionals to shape the evaluation infrastructure for responsible data and AI innovation for animals to avoid the field becoming fragmented by incompatible datasets, unvalidated tools, and opaque claims. The initiative's strength lies in its breadth, but its first success will depend on focus: one complete, useful, well-governed benchmark loop that demonstrates how veterinary relevance and technical reproducibility can work together.

The long-term aspiration remains to support the application of AI tools for the benefit of “any animal, any disease, any responsible task.” The near-term task is narrower and more urgent, scoped for proof of concept: build the first benchmark package, prove that stakeholders can use it, govern it responsibly, learn from the process, and expand only after building a credible foundation. If VETNET can accomplish this, it stands to become the trusted reference point for responsible veterinary AI innovation.

# Acknowledgments and Disclosures

*Funding*

The AI Thought Summit was funded by the Cornell AI Initiative at Cornell University to RI, PB, JJS, and CC. The efforts of author KK were funded by NIH T32ODO011000.

*Conflicts of Interest*

The authors declare no conflicts of interest.

*Disclosure*

The content is solely the responsibility of the authors and does not necessarily represent the official views of any government agency or private entity, including the National Center for Research Resources and the National Institutes of Health.

# Appendix A | Thought Summit Visual Summaries

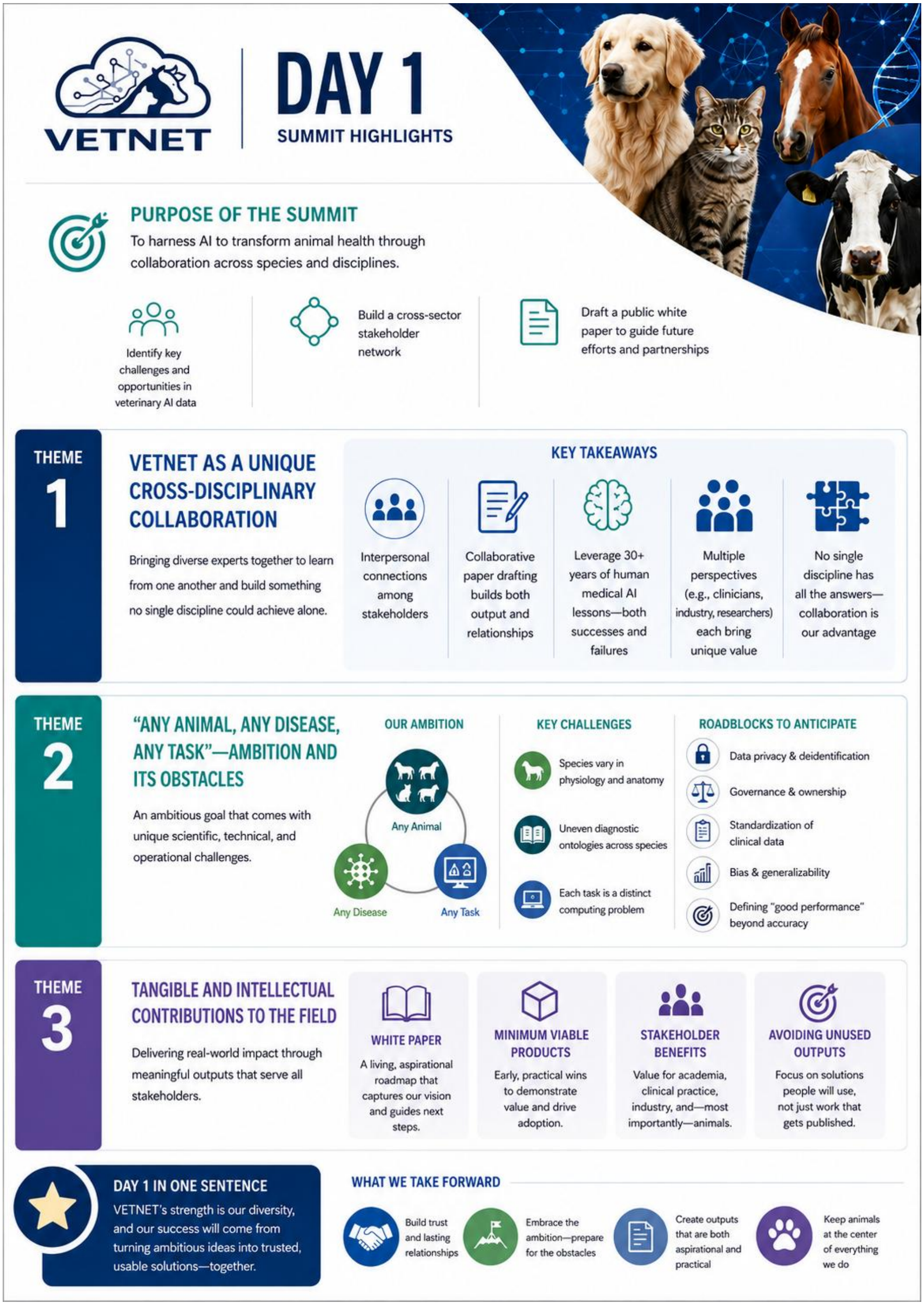

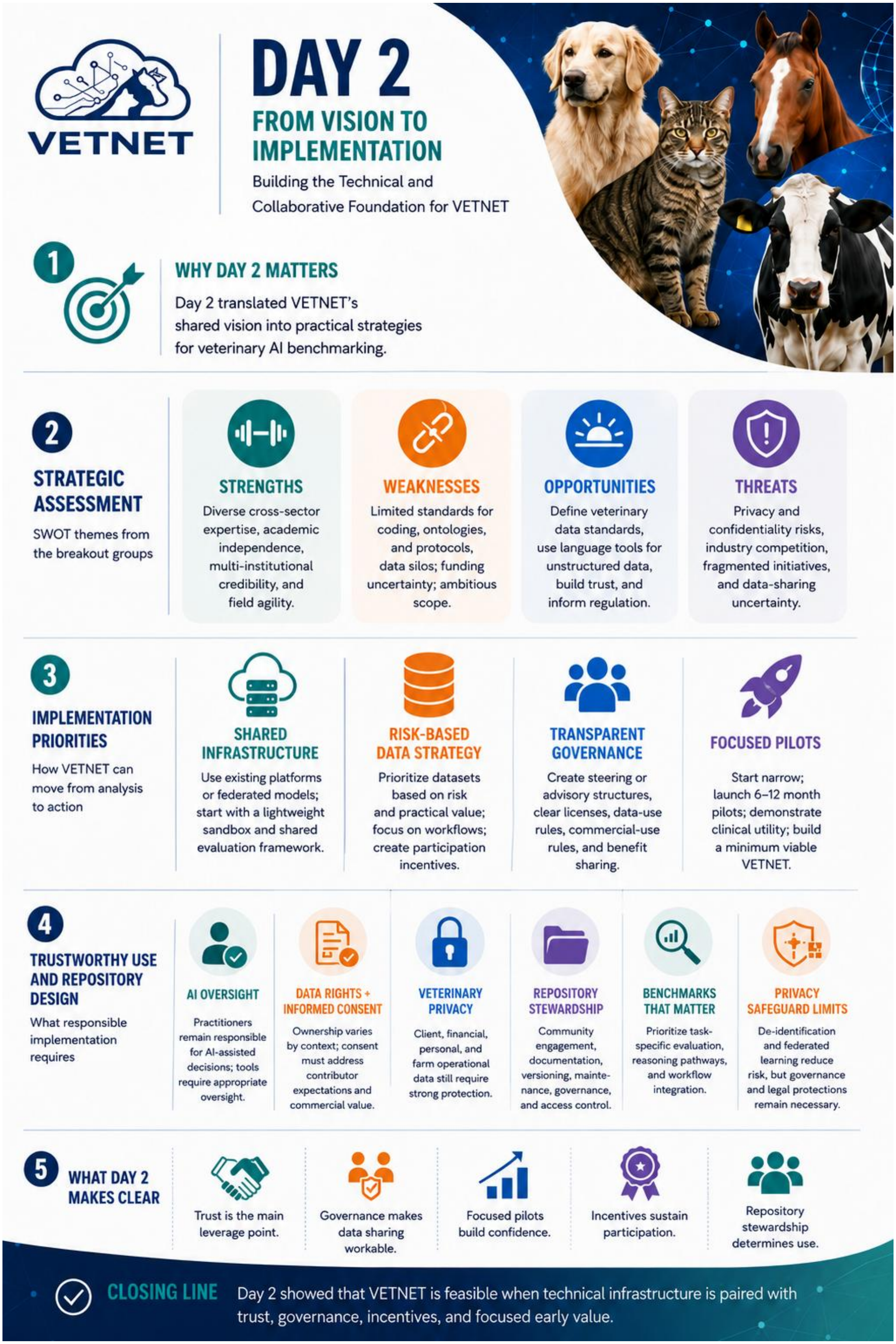
VETNET
DAY 2
FROM VISION TO IMPLEMENTATION
Building the Technical and Collaborative Foundation for VETNET
1
WHY DAY 2 MATTERS
Day 2 translated VETNET's shared vision into practical strategies for veterinary AI benchmarking.
2
STRATEGIC ASSESSMENT
SWOT themes from the breakout groups
STRENGTHS
Diverse cross-sector expertise, academic independence, multi-institutional credibility, and field agility.
WEAKNESSES
Limited standards for coding, ontologies, and protocols, data silos; funding uncertainty; ambitious scope.
OPPORTUNITIES
Define veterinary data standards, use language tools for unstructured data, build trust, and inform regulation.
THREATS
Privacy and confidentiality risks, industry competition, fragmented initiatives, and data-sharing uncertainty.
3
IMPLEMENTATION PRIORITIES
How VETNET can move from analysis to action
SHARED INFRASTRUCTURE
Use existing platforms or federated models; start with a lightweight sandbox and shared evaluation framework.
RISK-BASED DATA STRATEGY
Prioritize datasets based on risk and practical value; focus on workflows; create participation incentives.
TRANSPARENT GOVERNANCE
Create steering or advisory structures, clear licenses, data-use rules, commercial-use rules, and benefit sharing.
FOCUSED PILOTS
Start narrow; launch 6–12 month pilots; demonstrate clinical utility; build a minimum viable VETNET.
4
TRUSTWORTHY USE AND REPOSITORY DESIGN
What responsible implementation requires
AI OVERSIGHT
Practitioners remain responsible for AI-assisted decisions; tools require appropriate oversight.
DATA RIGHTS + INFORMED CONSENT
Ownership varies by context; consent must address contributor expectations and commercial value.
VETERINARY PRIVACY
Client, financial, personal, and farm operational data still require strong protection.
REPOSITORY STEWARDSHIP
Community engagement, documentation, versioning, mainte-nance, governance, and access control.
BENCHMARKS THAT MATTER
Prioritize task-specific evaluation, reasoning pathways, and workflow integration.
PRIVACY SAFEGUARD LIMITS
De-identification and federated learning reduce risk, but governance and legal protections remain necessary.
5
WHAT DAY 2 MAKES CLEAR
Trust is the main leverage point.
Governance makes data sharing workable.
Focused pilots build confidence.
Incentives sustain participation.
Repository stewardship determines use.
CLOSING LINE
Day 2 showed that VETNET is feasible when technical infrastructure is paired with trust, governance, incentives, and focused early value.

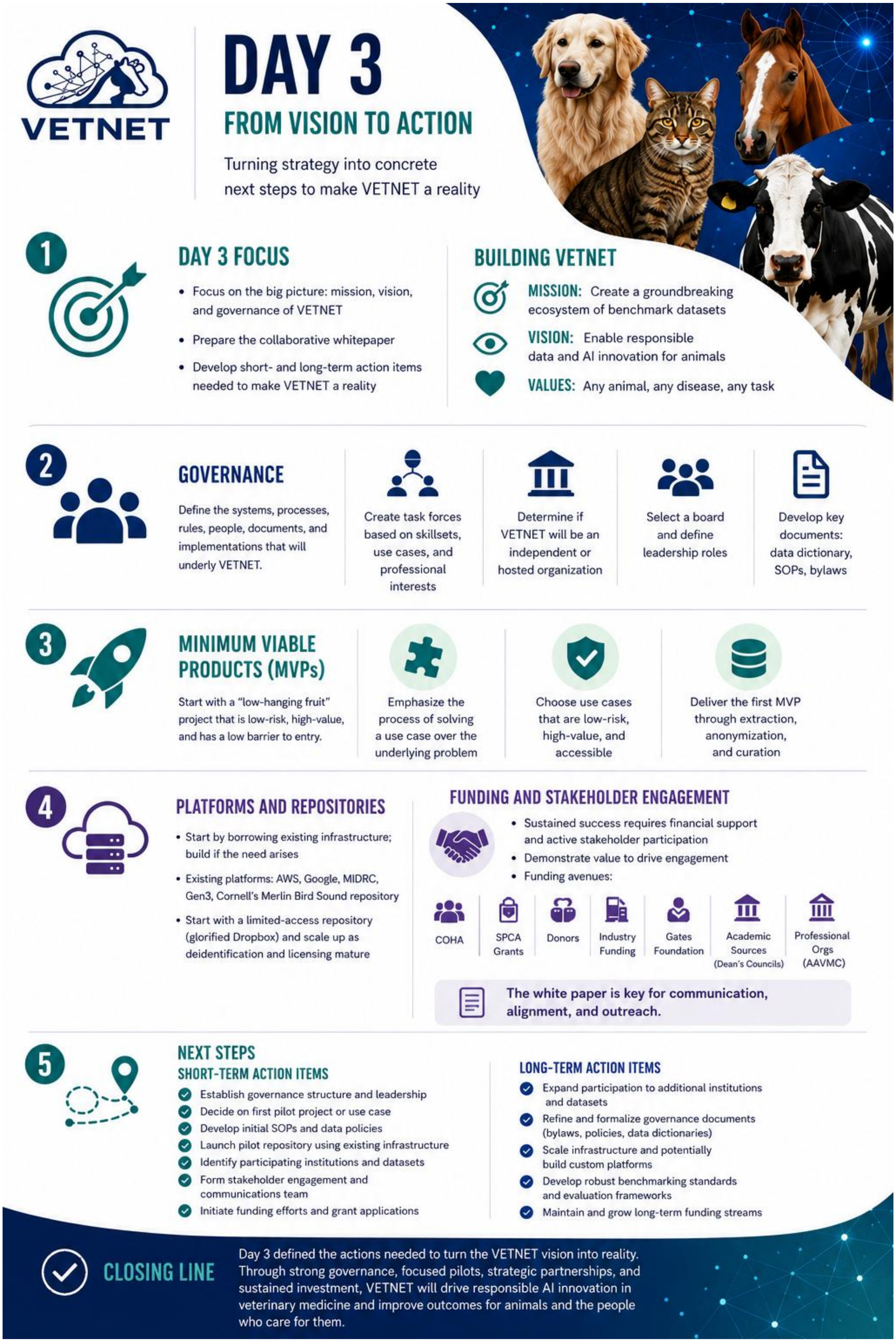

VETNET
DAY 3
FROM VISION TO ACTION
Turning strategy into concrete next steps to make VETNET a reality
1
DAY 3 FOCUS
• Focus on the big picture: mission, vision, and governance of VETNET
• Prepare the collaborative whitepaper
• Develop short- and long-term action items needed to make VETNET a reality
BUILDING VETNET
MISSION: Create a groundbreaking ecosystem of benchmark datasets
VISION: Enable responsible data and AI innovation for animals
VALUES: Any animal, any disease, any task
2
GOVERNANCE
Define the systems, processes, rules, people, documents, and implementations that will underly VETNET.
Create task forces based on skillsets, use cases, and professional interests
Determine if VETNET will be an independent or hosted organization
Select a board and define leadership roles
Develop key documents: data dictionary, SOPs, bylaws
3
MINIMUM VIABLE PRODUCTS (MVPs)
Start with a "low-hanging fruit" project that is low-risk, high-value, and has a low barrier to entry.
Emphasize the process of solving a use case over the underlying problem
Choose use cases that are low-risk, high-value, and accessible
Deliver the first MVP through extraction, anonymization, and curation
4
PLATFORMS AND REPOSITORIES
• Start by borrowing existing infrastructure; build if the need arises
• Existing platforms: AWS, Google, MIDRC, Gen3, Cornell's Merlin Bird Sound repository
• Start with a limited-access repository (glorified Dropbox) and scale up as deidentification and licensing mature
FUNDING AND STAKEHOLDER ENGAGEMENT
• Sustained success requires financial support and active stakeholder participation
• Demonstrate value to drive engagement
• Funding avenues:
COHA
SPCA Grants
Donors
Industry Funding
Gates Foundation
Academic Sources (Dean's Councils)
Professional Orgs (AAVMC)
The white paper is key for communication, alignment, and outreach.
5
NEXT STEPS
SHORT-TERM ACTION ITEMS
Establish governance structure and leadership
Decide on first pilot project or use case
Develop initial SOPs and data policies
Launch pilot repository using existing infrastructure
Identify participating institutions and datasets
Form stakeholder engagement and communications team
Initiate funding efforts and grant applications
LONG-TERM ACTION ITEMS
Expand participation to additional institutions and datasets
Refine and formalize governance documents (bylaws, policies, data dictionaries)
Scale infrastructure and potentially build custom platforms
Develop robust benchmarking standards and evaluation frameworks
Maintain and grow long-term funding streams
CLOSING LINE
Day 3 defined the actions needed to turn the VETNET vision into reality. Through strong governance, focused pilots, strategic partnerships, and sustained investment, VETNET will drive responsible AI innovation in veterinary medicine and improve outcomes for animals and the people who care for them.

## Appendix B | Minimum Dataset Documentation Template

- Dataset title, version, date, contributors, and contact or governance point of contact
- Species, breeds or populations represented, practice settings, geography, modality, collection timeframe, and inclusion/exclusion criteria
- Raw data description, processed data description, cleaning steps, deidentification protocol, transformation scripts, and known data losses
- Label definition, labeling protocol, labeler expertise, disagreement adjudication, uncertainty representation, and protected holdout rules
- Metadata fields, missingness, device/vendor details, practice-management-system export details, and terminology or ontology mappings
- Access model, license, permitted uses, prohibited uses, commercial-use rules, attribution language, and withdrawal/correction procedure
- Known limitations, anticipated biases, subgroup gaps, clinical-risk warnings, equity/access limitations, and interpretation guidance
- External validation, temporal validation, subgroup reporting, artifact/shortcut probes, and drift-monitoring plans where feasible

## Appendix C | Tiered Access Model

| Access tier | Appropriate for | Example controls |
|---|---|---|
| Open release | Low governance-risk data with strong deidentification, consent, documentation, and low commercial sensitivity. | Public download, citation requirements, clear disclaimers, versioned releases. |
| Registered access | Moderate-risk data requiring user identity, agreement to terms, and basic monitoring. | User registration, data-use agreement, access logs, and revocable credentials. |
| Controlled access | Sensitive clinical, farm, proprietary, or high-reidentification-risk data. | Application review, institutional approval, audited access, restricted redistribution, and commercial use limits. |
| Federated or hosted evaluation | Data that cannot move because of privacy, commercial sensitivity, cross-border restrictions, or contributor preference. | Models or submissions move to data; protected outputs; differential privacy or additional safeguards where appropriate. |
| Low-resource participation support | Benchmarks intended for global or small-practice participation. | Lightweight baselines, documented compute needs, simplified submissions, cloud credits where available, and nonexclusive access pathways. |

## Appendix D | Glossary for Cross-Sector Readers

| Term | Working meaning in this paper |
|---|---|
| Benchmark | A defined evaluation package for AI tools or models that includes task, data or access, labels, metrics, code, documentation, governance, and interpretation rules |
| Evaluation harness | Runnable code and file formats that let users submit predictions and compute standardized scores |
| Holdout set | A protected test set whose labels are not exposed to model developers, used to reduce overfitting |
| External validation | Evaluation on data from different sites, practices, equipment, geographies, or institutions than the training data |
| Temporal validation | Evaluation on data from a later time period than the training data to test robustness to change over time |
| Shortcut learning | A model learns an easy but misleading correlate, such as device or site artifacts, instead of the intended clinical signal |
| Distribution shift | The deployment data differ from the development data, often reducing performance |
| Calibration | How well predicted confidence matches observed correctness |
| Automation bias | Humans over-rely on automated outputs even when independent evidence suggests the output may be wrong |
| Deskilling | Erosion of human expertise or judgment through excessive reliance on automation |
| Federated evaluation | Data remain with the holder; models or submissions are evaluated against data without broadly releasing raw records |
| Data card/model card | Structured documentation describing data or model provenance, limitations, intended uses, and risks |
| Croissant | A structured metadata vocabulary for ML datasets that supports discoverability, interoperability, and reproducibility |
| Inner-loop evaluation | Offline benchmark evaluation under controlled conditions |
| Outer-loop validation | Prospective, workflow, safety, welfare, ethical, legal, and deployment evidence needed before real-world use claims |